\documentclass{ieeeaccess}
\usepackage{cite}
\usepackage{amsmath,amssymb,amsfonts}
\usepackage{algorithmic}
\usepackage{graphicx}
\usepackage{textcomp}

\usepackage{booktabs}
\usepackage{array}
\usepackage{url}
\usepackage[hidelinks]{hyperref}
\usepackage{cleveref}
\usepackage{multirow}
\usepackage{adjustbox}
\usepackage{amsmath}
\usepackage{caption}
\usepackage{tabularx}  
\usepackage{xurl}     
\usepackage{hyphenat} 
\usepackage{soul} 
\usepackage{makecell} 
\usepackage{threeparttable}
\sethlcolor{yellow}

\newcommand{\huggingfacemap}[1]{\href{https://huggingface.co/datasets/#1}{\texttt{#1}}} 

\usepackage{bm}
\makeatletter
\AtBeginDocument{\DeclareMathVersion{bold}
\SetSymbolFont{operators}{bold}{T1}{times}{b}{n}
\SetSymbolFont{NewLetters}{bold}{T1}{times}{b}{it}
\SetMathAlphabet{\mathrm}{bold}{T1}{times}{b}{n}
\SetMathAlphabet{\mathit}{bold}{T1}{times}{b}{it}
\SetMathAlphabet{\mathbf}{bold}{T1}{times}{b}{n}
\SetMathAlphabet{\mathtt}{bold}{OT1}{pcr}{b}{n}
\SetSymbolFont{symbols}{bold}{OMS}{cmsy}{b}{n}
\renewcommand\boldmath{\@nomath\boldmath\mathversion{bold}}}
\makeatother

\def\BibTeX{{\rm B\kern-.05em{\sc i\kern-.025em b}\kern-.08em
    T\kern-.1667em\lower.7ex\hbox{E}\kern-.125emX}}

\begin{document}
\history{Received 10 July 2025, accepted 30 August 2025, date of publication 2 September 2025, date of current version 8 September 2025.}
\doi{10.1109/ACCESS.2025.3605290}

\title{Entropy-Based Data Selection for\\Language Models}
\author{\uppercase{Hongming Li}\authorrefmark{1},
\uppercase{Yang Liu}\authorrefmark{2}, 
\uppercase{and Chao Huang}\authorrefmark{3}}

\address[1]{School of Computer and Communication Engineering, University of Science and Technology Beijing, Beijing 100083, China}
\address[2]{National Key Laboratory of General Artificial Intelligence, Beijing Institute for General Artificial Intelligence, Beijing 100086, China}
\address[3]{Shunde Innovation School, University of Science and Technology Beijing, Foshan 528399, China}

\tfootnote{This work was supported in part by the National Natural Science Foundation of China under Grant 62372039 and Grant 62002016, and in part by the Guangdong Basic and Applied BasicmResearch Foundation under Grant 2022A1515240044.}

\markboth
{H. Li \headeretal: Entropy-Based Data Selection for Language Models}
{H. Li \headeretal: Entropy-Based Data Selection for Language Models}

\corresp{Corresponding author: Chao Huang (chaohuang@ustb.edu.cn).}

\begin{abstract}
Modern language models (LMs) increasingly require two critical resources: computational resources and data resources. Data selection techniques can effectively reduce the amount of training data required for fine-tuning LMs. However, their effectiveness is closely related to computational resources, which always require a high compute budget. Owing to the resource limitations in practical fine-tuning scenario, we systematically reveal the relationship between data selection and uncertainty estimation of selected data. Although large language models (LLMs) exhibit exceptional capabilities in language understanding and generation, which provide new ways to alleviate data scarcity, evaluating data usability remains a challenging task. This makes efficient data selection indispensable. To mitigate these issues, we propose \textit{Entropy-Based Unsupervised Data Selection (EUDS)} framework. Empirical experiments on sentiment analysis (SA), topic classification (Topic-CLS), and question answering (Q\&A) tasks validate its effectiveness. EUDS establishes a computationally efficient data-filtering mechanism. Theoretical analysis and experimental results confirm the effectiveness of our approach. EUDS significantly reduces computational costs and improves training time efficiency with less data requirement. This provides an innovative solution for the efficient fine-tuning of LMs in the compute-constrained scenarios.
\end{abstract}

\begin{keywords}
Uncertainty estimation, language models, synthetic data, data selection.
\end{keywords}

\titlepgskip=-21pt

\maketitle

\section{Introduction}
\label{Introduction}
\PARstart{A}{s} frontier progress continues and the need for handling increasingly complex language tasks rises, language models (LMs) have achieved impressive human-like performance for addressing a wide array of language understanding and generation tasks. 

The rapid development of LMs, particularly large language models (LLMs), has introduced significant resource demands, primarily in terms of computational and data resources. 
Training these models incurs exceedingly high computational expenses. In many cases, the total computational budget was predetermined. Although resource requirements during the fine-tuning phase are typically lower than those during pre-training, fine-tuning can still pose substantial computational challenges for certain tasks. In scenarios with limited computational resources, the efficient allocation of these resources is critically important.
Data selection techniques create optimal subsets based on specific objective functions from the original dataset. These techniques reduce computational demands by minimizing the training samples required for pre-training or fine-tuning \cite{John01021975, yin2024compute}.

\begin{figure*}[t!]
\centering
\includegraphics[width=\textwidth]{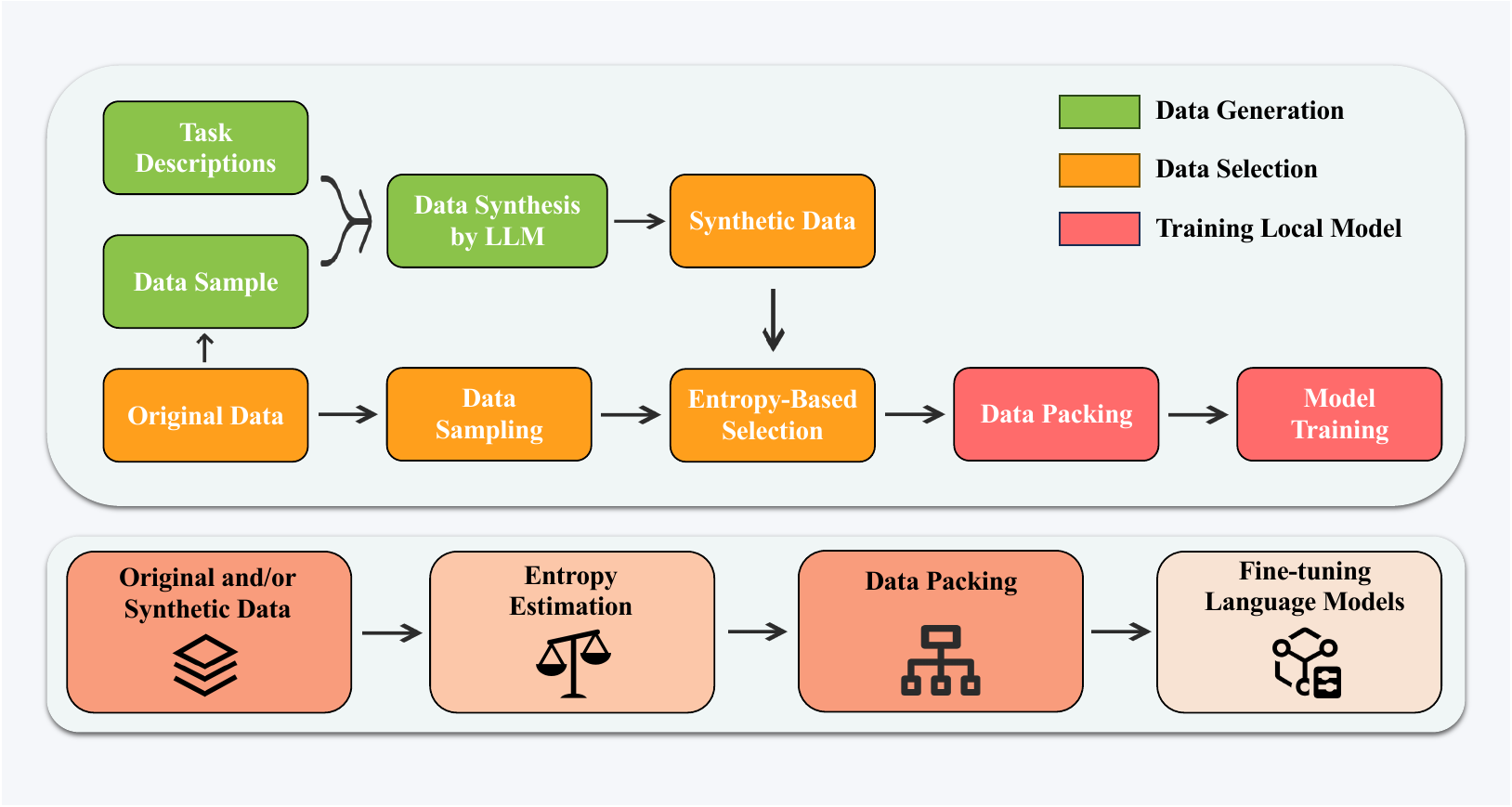}
\caption{Overview of our framework.}
\label{fig:Overview of our framework}
\end{figure*}

Fine-tuning is typically required to adapt an LM to a specific task or domain \cite{agiza2024polituneanalyzingimpactdata}. The emergence of LLMs has shifted data-centric development from maximizing volume to strategically selecting high-value data. This shift, driven by the high computational cost of fine-tuning, has inspired a range of efficient data selection approaches that go beyond traditional heuristics or proxy-based methods. Recent trends primarily include influence-based techniques, which estimate the importance of individual samples to model alignment \cite{nikdan2025efficientdataselectionscale, anonymous2025ideal}, and uncertainty-driven strategies \cite{ liu2025selectitselectiveinstructiontuning} as well as model-oriented methods\cite{du2023modsmodelorienteddataselection} that exploit LLM-internal signals to guide selection without external proxies. Other efforts, such as speculative coreset construction \cite{zhang2025staff} and self-reflective filtering via LLMs themselves \cite{peng2025datamandatamanagerpretraining}, further advance the paradigm toward adaptive, efficient, and architecture-aware data curation. 

Constructing a robust LM relies heavily on large volumes of high-quality data \cite{gandhi-etal-2024-better, long-etal-2024-llms}. Data scarcity, privacy concerns, and insufficient diversity are significant challenges. Relying solely on human-generated data is highly problematic \cite{liu2024bestpracticeslessonslearned}. This approach often leads to limited usable data. Current LLMs have exceptional generative capabilities. Leveraging LLMs for human-generated data has emerged as an effective solution to address this issue \cite{xu-etal-2020-megatron, ding-etal-2023-gpt, long-etal-2024-llms}. However, LLMs may encounter limitations in generating data that fully capture the breadth and complexity of real-world language usage, resulting in uneven synthetic data quality \cite{li-etal-2023-synthetic}.  Therefore, even when synthetic data are used to mitigate data resource constraints, data selection remains an indispensable component of the process.

Although various existing data selection methods have reduced the training data volume to some extent \cite{bai2022constitutionalaiharmlessnessai, touvron2023llama2openfoundation, kim2024prometheusinducingfinegrainedevaluation}, these approaches still exhibit notable limitations. In particular, under conditions of limited computational resources, traditional data selection methods often struggle to balance efficiency and effectiveness. Consequently, there is an urgent need for the introduction of novel data selection techniques that can enable more efficient data screening in computationally constrained environments.

We propose an Entropy-Based Unsupervised Data Selection (EUDS) framework, specifically designed for the optimal selection of original and synthetic data. This framework employs three entropy-based evaluation mechanisms for data selection: Information Entropy (IE) at the lexical level, Generative Entropy (GE) based on perplexity, and Semantic Entropy (SE) at the semantic level \cite{Farquhar2024, tjandra2024finetuninglargelanguagemodels}

Notably, the EUDS framework operates as an isolated preprocessing step prior to LM fine-tuning. Rather than being integrated into the internal architecture of a LM, EUDS filters both original and synthetic data based on entropy scores before training. This design allows EUDS to remain model-agnostic and readily applicable to a wide range of LMs, including but not limited to BERT and GPT-based models.

Our core contributions are as follows: 
\begin{itemize}     
   \item[$\bullet$] \textbf{Innovative Framework Construction}: A simple yet efficient EUDS framework is proposed, which is designed with a three-tier mechanism–IE–GE–SE —for data selection. Through this approach, the risk of model failure is mitigated, and the performance of downstream classification tasks is significantly enhanced.     
   \item[$\bullet$] \textbf{Verification in Computationally Constrained Scenarios}: Under conditions of limited computational resources, the framework’s selected original/synthetic data can independently support model training. Despite a significant reduction in data volume, it can maintain or even improve performance metrics for downstream tasks.     
   \item[$\bullet$] \textbf{Optimization of Data Mixing Strategy}: An original and synthetic data augmentation strategy is constructed in response to data resource constraints. The end-to-end classification performance is enhanced through the synergistic combination of synthetic data selected using the EUDS framework with the original data. By further incorporating it into the selected original data, the model performance can be maintained while the data requirements can be further reduced.
   \item[$\bullet$] \textbf{Expansion of the Entropy Combination Strategy}: An optimization strategy based on combined entropy has been developed. Through collaborative evaluation across different entropy measures, the demands on computational and data resources are further reduced.    
   \item[$\bullet$] \textbf{Large-Scale Validation Experiment}: An extensibility verification study of the single-entropy evaluation mechanism is conducted within the framework developed in this study. Through the results, the EUDS framework is confirmed to maintain baseline performance levels even in large-scale data scenarios, thereby validating scalability of the proposed method.
\end{itemize}

\section{Related Work}\label{sec:related_work}
\subsection{Pre-trained Language Models}

 Pre-trained language models (PLMs) have catalyzed a paradigm shift in Natural Language Processing (NLP).  Built on the Transformer architecture, PLMs effectively capture deep semantic representations of language through self-supervised learning mechanisms, such as BERT’s masked language model \cite{devlin-etal-2019-bert}. This enables the establishment of a two-stage universal framework of pre-training and fine-tuning. In this framework, the model is first pretrained on large-scale corpora to capture universal language patterns \cite{JMLR:v21:20-074, RATHNAYAKE2024107239, lu-etal-2024-mitigating}. It is then fine-tuned for specific downstream tasks to adapt to domain-specific features \cite{hu2022lora, Verma2024, vijayaraghavan-etal-2024-self, villa2024adaptive}. With the scaling of model size and optimization of training strategies, PLMs have become a core technology for handling various downstream NLP tasks \cite{ho2024surveypretrainedlanguagemodels}. Our main research method is to fine-tune LMs to better adapt them to downstream text classification tasks.

\subsection{Synthetic Data}
To address the issue of limited data resources, researchers have turned to LLMs to generate synthetic data, thereby enabling efficient data augmentation. Studies have shown that synthetic data generated by LLMs can effectively enhance a model's adaptability and robustness in specific tasks \cite{kumar-etal-2020-data, yoo-etal-2021-gpt3mix-leveraging, hartvigsen-etal-2022-toxigen}. Focusing on text classification tasks, research has delved into the complex mechanisms through which the conditions for synthetic data generation influence the model training. Li et al. \cite{li-etal-2023-synthetic} demonstrates through systematic experiments that task subjectivity serves as a key factor in moderating the effectiveness of synthetic data. Furthermore, the structured generation process developed by Wang et al.\cite{wang-etal-2023-lets} successfully narrows the representation gap between synthetic and real data. These groundbreaking advancements not only break through traditional data acquisition bottlenecks but also significantly enhance the model's transferability and resilience to disturbances by generating diverse training samples \cite{sahu-etal-2022-data}. We use synthetic data to mitigate the issue of limited data resources and verify the effectiveness of our data selection method on such data.

\subsection{Uncertainty Estimation}
Uncertainty estimation methods have been widely explored in the multitask domain. Fomicheva et al. \cite{fomicheva2020unsupervised} quantify model uncertainty using various techniques, thereby improving the accuracy and reliability of unsupervised quality estimation. Building on uncertainty quantification methods, Desai et al. and Jiang et al. \cite{desai-durrett-2020-calibration, jiang-etal-2021-know} propose calibration techniques for text classification tasks. Xiao et al.\cite{xiao-wang-2021-hallucination} empirically points out that measuring model uncertainty can effectively enhance task performance for tasks such as sentiment analysis and named entity recognition. In Natural Language Generation (NLG) tasks, the uncertainty estimation of model outputs is a critical issue. Traditional uncertainty estimation methods, such as predictive entropy, often overlook semantic equivalence, which leads to inaccurate estimates. To address this, ``semantic entropy" is used to improve the uncertainty estimation.  Clustering algorithms group semantically equivalent sentences into categories. The entropy of these categories is computed for a more accurate uncertainty estimate \cite{kuhn2023semantic, nikitin2024kernel, Farquhar2024}.  This study innovatively apply the semantic entropy framework in reverse to assess the original data. This reverse transfer not only extends the application boundaries of semantic entropy but also establishes a unified evaluation system for both generated and original data in terms of the information metrics.

\subsection{Data Selection}
\label{subsec:Data Selection}
Under fixed computational budgets and resource constraints, directly training or fine-tuning all usable textual data may not be the most optimal or feasible strategy. Given the significant variations in the quality, information richness, and diversity of usable data, the selection of training data plays a decisive role in LM performance \cite{A_Survey_on_Data_Selection_for_Language_Models}. The goal of data selection is to choose a subset from the given training data such that the model can achieve a performance comparable to that of the full dataset while adhering to computational resource limitations.

Proxy-based data selection methods are typically computationally expensive and are primarily suitable for selection from human-curated pretraining data mixtures \cite{grattafiori2024llama3herdmodels, li2024datacomplmsearchgenerationtraining}. 
Furthermore, the use of LLMs to allocate utilities has been explored. One approach involves the model's forward inference, such as using various utility scores and perplexity \cite{marion2023moreinvestigatingdatapruning, wettig2024quratingselectinghighqualitydata, yin2024compute}. These methods provide task-agnostic heuristics that often correlate with informativeness or difficulty.

Another stream of research employs gradient-based influence estimation to assess the impact of individual training samples on model parameters or downstream performance. Influence function-style selection has been widely adopted for this purpose \cite{pmlr-v139-killamsetty21a, 10.5555/3600270.3602281, 10.5555/3692070.3694291}, offering a more fine-grained evaluation of training data utility, albeit at the cost of increased computation.

Beyond classical approaches, recent research emphasizes adaptive and architecture-aware strategies that directly leverage model-internal signals to guide selection, moving beyond proxy-based heuristics. This paradigm shift has inspired a range of techniques—spanning influence-based, uncertainty-driven, speculative scoring, and LLM-internal self-reflection approaches—enabling more scalable, task-adaptive data curation in LLMs \cite{nikdan2025efficientdataselectionscale, anonymous2025ideal, zhang2025staff, peng2025datamandatamanagerpretraining}. Similar trends are also observed in other modalities, such as sign language recognition, where selective input filtering has significantly improved model robustness and efficiency \cite{10538314, 10216946, ABDULLAHI2024123258}.

\section{Method}\label{sec:methodology}
\subsection{Synthetic Data Generation Method}
\label{subsec:data integration}
LLMs have demonstrated remarkable capabilities in language understanding and generation tasks, providing new approaches to address the challenges of data acquisition. Building on this, we utilize GPT-4o to generate large volumes of synthetic data along with the corresponding labels \cite{ding-etal-2023-gpt}, which are used to train the local models. This method requires only a small number of examples to generate a complete training dataset, effectively alleviating the data scarcity issue in specific tasks or domains.

In generating synthetic data, GPT-4o creates the required content through customized instructions and prompt texts \cite{JMLR:v25:23-0870}. Leveraging this feature, we design a systematic framework for generating synthetic data.  
Specific prompt templates are designed for different text classification tasks, including sentiment analysis (SA), topic classification (Topic-CLS), and question answering (Q\&A) tasks. Few-shot prompting techniques were used to guide the GPT model in generating the synthetic data. Specifically, we randomly sample from the original dataset of a particular task or domain, embedding the samples into predefined prompt templates. By adjusting the temperature parameter, diverse content is generated to align with specific requirements, resulting in synthetic data tailored to the task \cite{lupidi2024source2synth, Zhao2025}. A closed-loop process from ``original data guidance'' to ``synthetic data augmentation'' is successfully implemented using this framework.

\subsection{Principle of Data Selection}
\label{subsec:data selection}
The EUDS approach is proposed, which is designed to efficiently select original and synthetic data, thereby reducing the data volume required for fine-tuning and overcoming. In this study, we focused on three types of entropy and evaluated the effectiveness of both individual and combined entropy measures in the context of data selection.

\textbf{IE}: Based on classical Shannon Entropy, denoted as $\mathrm{H}(\cdot)$, this method estimates the level of uncertainty in the probability distributions of various linguistic units across texts. The calculation formula is as follows:
\begin{equation}
    \mathrm{H}(x_i) = -\sum_{i=1}^n \mathrm{P}\left(x_i\right) \log _2 \mathrm{P}\left(x_i\right)
\end{equation}
Here, \(\mathrm{P}(x_i)\) represents the probability of the \( i \)-th linguistic unit occurring in the text,and the summation covers all possible \( n \) linguistic units.

Linguistic units of different granularities, such as unigrams, bigrams, and trigrams, are combined. Their weighted contributions are integrated to quantify the diversity and uncertainty of the text, leading to the definition of IE as:
\begin{equation}
    \mathrm{IE}(x) = \sum_{n=1}^{3} \alpha_n \sum_{x_i \in \mathcal{V}_{\mathrm{n\text{-gram}}}} \mathrm{P}_{\mathrm{n\text{-gram}}}(x_i) \log_2 \mathrm{P}_{\mathrm{n\text{-gram}}}(x_i)
\end{equation}
Here, \( \mathrm{P_{n\text{-gram}}}(x_i) \) is the probability of the \( i \)-th word in the \( \mathrm{n} \)-gram, defined as the relative frequency of that n-gram in the text; \( \mathcal{V}_\mathrm{{n\text{-gram}}} \) is the set of all possible \( \mathrm{n} \)-gram; and \( \alpha_n \) is the weight coefficient used to adjust the contribution of different \( \mathrm{n} \)-gram to the overall entropy.

\textbf{GE}: This entropy measure quantifies the predictive uncertainty during text generation and assesses the difficulty of the model in predicting the next word.
\begin{equation}
    \mathrm{GE}(x) = - \frac{1}{n} \sum_{i=1}^n \log_2 \mathrm{P}(s_i \mid s_{<i}, x)
\end{equation}
Here, \(\mathrm{P}(s_i\ |\ s_{<i},x)\) represents the probability of the model predicting the \( i \)-th word given the context \( s_{<i} \).

\textbf{SE}: By aggregating semantically equivalent generation outcomes (i.e., different expressions conveying the same meaning), we measure uncertainty at the semantic level, thereby avoiding overestimation caused by expression diversity.
\begin{equation}
\mathrm{SE}(x) = -\sum_{i=1}^{|C|} \mathrm{P}(C_i|x)\log \mathrm{P}(C_i|x)
\end{equation}
Here, \( C \) denotes the set of semantic equivalence classes, and \(\mathrm{P}(C_i|x)\) represents the probability that the generated outcome for input \( x \) belongs to the 
\( i \)-th semantic class.

\begin{figure*}[t!]{}
\centering
\includegraphics[width=1.0\textwidth]{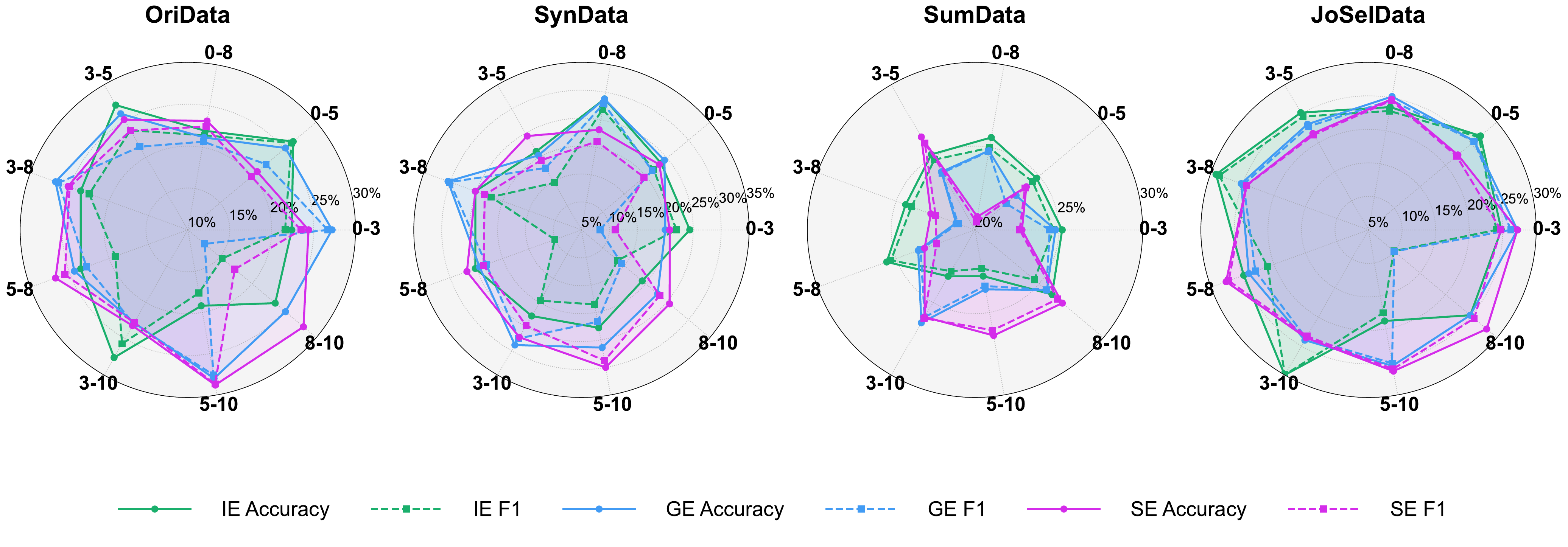}
\caption{The impact of different entropy intervals on model performance.}
\label{fig:sensitive_interval}
\end{figure*}

\subsection{Entropy-based selection workflow and methodology}
\label{subsec:clarification}

The selection process consists of the following key stages:

\paragraph{Entropy Computation}
For each candidate sample $x$, the values of $\mathrm{IE}(x)$, $\mathrm{GE}(x)$, and $\mathrm{SE}(x)$ are computed. These entropy metrics respectively reflect lexical diversity, predictive difficulty during token generation, and semantic ambiguity.

\paragraph{Interval-Based Selection via Subset Generalization}
Rather than assuming that samples with low or high entropy are inherently better or worse, we adopt a data-driven strategy to determine the optimal entropy range. Specifically, we randomly sample a representative subset of the full candidate dataset. On this subset, we compute entropy scores and divide them into multiple quantile-based intervals. Each interval is then evaluated by training a model on the corresponding data and measuring downstream validation performance.

The optimal entropy interval of this subset—defined as the interval(s) that achieve a favorable balance between data reduction and accuracy gain, and which may vary depending on the desired trade-off—is essentially identical to that of the complete candidate dataset. We then apply this interval to the full dataset to select samples for fine-tuning. This method ensures that the entropy-performance relationship observed in the representative subset generalizes to the full dataset, enabling scalable and effective data selection without exhaustive global search.

\paragraph{Integration with LM Fine-Tuning}
The entropy-based selection is applied as a preprocessing step before model fine-tuning. The selected samples are directly used to construct the fine-tuning dataset. This static selection design ensures that the training dynamics of the LM are not altered and no modification to optimization objectives or architectures is required.

\paragraph{Computational Complexity}
The computational costs of the entropy calculations are as follows:
\begin{itemize}
    \item \textbf{IE}: Computed via $n$-gram frequency counts; complexity is $O(N)$, where $N$ is the number of tokens.
    \item \textbf{GE}: Requires forward passes through the model during autoregressive generation; complexity is $O(N \cdot T)$, where $T$ is the average decoding length.
    \item \textbf{SE}: Involves sampling diverse generations and clustering them semantically; complexity is approximately $O(N \cdot C)$, with $C$ being the number of sampled generations per input.
\end{itemize}

\paragraph{Sensitive Parameters}

The performance of the EUDS framework is affected by the choice of entropy interval (Figure \ref{fig:sensitive_interval}). Experimental results show that the entropy interval is the most sensitive parameter influencing model performance. Under the same entropy, different entropy intervals lead to significant fluctuations in model performance, and this phenomenon exists across different data sources.

\begin{table}[ht]
\caption{Entropy Level Classification with Interval Ranges.}
\small
\renewcommand{\arraystretch}{0.3}
\centering
\resizebox{\columnwidth}{!}{
\begin{tabular}{>{\bfseries}ccc}
\toprule
\textbf{Entropy Level} & \textbf{Numerical Interval} & \textbf{Abbreviation} \\
\midrule
Low Entropy     & $(0.0, 3.0)$; $(3.0, 5.0)$; $(0.0, 5.0)$      & 0-3; 3-5; 0-5      \\
Medium Entropy  & $(0.0, 8.0)$; $(3.0, 10.0)$; $(3.0, 8.0)$     & 0-8; 3-10; 3-8     \\
High Entropy    & $(5.0, 8.0)$; $(8.0, 10.0)$; $(5.0, 10.0)$    & 5-8; 8-10; 5-10    \\
\bottomrule
\end{tabular}
}
\vspace{0.2cm}
\label{tab:entropy_intervals}
\end{table}

Based on the estimated results, the entropy values of both original and synthetic data are normalized and mapped to a scoring range of 0–10. The data is dynamically partitioned by the EUDS model into several non-uniform intervals based on the distribution characteristics of the entropy values (refer to Table \ref{tab:entropy_intervals} for interval definitions and Table \ref{tab:entropy_subset_summary} for resulting subset sizes). The effectiveness of this entropy-based selection strategy is systematically validated by comparing model performance after fine-tuning with data from each interval.

\begin{table}[ht]
\centering
\caption{Selected Subset Sizes for Representative Datasets Under Different Entropy Types and Data Configurations. Each Case Uses the Best-Performing Entropy Interval.}
\label{tab:entropy_subset_summary}
\normalsize
\setlength{\tabcolsep}{4pt}
\renewcommand{\arraystretch}{1.2}
\begin{threeparttable}
\begin{tabular}{ccccc}
\toprule
\makecell[c]{Dataset \\ (Entropy Type)} & \makecell[c]{Data \\ Type} & \makecell[c]{Entropy \\ Interval} & \makecell[c]{Original \\ Count} & \makecell[c]{Selected \\ Count} \\
\midrule
\multirow{4}{*}{\makecell[c]{MMLU \\ (IE)}} 
  & OriData   & 3–10         & 2000 & 1038 \\
  & SynData   & 3–8        & 2000 & 669  \\
  & SumData   & 8–10  & 4000 & 2013 \\
  & JoSelData & 3–10  & 4000 & 1720 \\
\midrule
\multirow{4}{*}{\makecell[c]{MMLU \\ (GE)}} 
  & OriData   & 5–10         & 2000 & 1200 \\
  & SynData   & 5–10        & 2000 & 1170  \\
  & SumData   & 8–10   & 4000 & 2029 \\
  & JoSelData & 0–5   & 4000 & 1630 \\
\midrule
\multirow{4}{*}{\makecell[c]{MMLU \\ (SE)}} 
  & OriData   & 5–10         & 2000 & 945  \\
  & SynData   & 5–10         & 2000 & 1881  \\
  & SumData   & 8–10    & 4000 & 3391 \\
  & JoSelData & 8–10    & 4000 & 1411  \\
\bottomrule
\end{tabular}
\begin{tablenotes}
    \small
    \item \textit{Note:} Each dataset comprises nine subsets (as shown in Table \ref{tab:entropy_intervals}). Their sizes are dynamically partitioned by the EUDS model based on entropy values, resulting in non-uniform subset sizes. Taking the MMLU dataset as an example, the table above shows the scale of its subsets after data selection when optimal performance is achieved under three entropy conditions.
\end{tablenotes}
\end{threeparttable}
\end{table}

\subsection{Downstream Tasks Training}
To validate the effectiveness of the EUDS framework in downstream tasks, a fine-tuning experimental paradigm is employed based on the BERT-base-uncased (BERT) model \cite{devlin-etal-2019-bert}. The self-attention mechanism of the Transformer architecture is fully utilized in this design. Through the dynamic capture of global dependencies between inputs and outputs, BERT is enabled to flexibly adapt to various downstream tasks. The specific implementation process includes the following core steps: 

\textbf{Data Preparation}: The original data from Section \ref{subsec:Original Datasets} and the generated data from Section \ref{subsec:Synthetic Datasets} are integrated, and after being selected using the method described in Section \ref{subsec:Data Selection}, they form the training set.

\textbf{Model Adaptation}: A task-specific classification head is added to the BERT model, where the hidden state of the [CLS] token is mapped to the task label space through learnable parameters.

\textbf{Optimal Allocation}: The model is fine-tuned using the hyperparameters mentioned in Section \ref{subsec:Technical Details}. During the training process, the loss values and performance metrics are monitored, and the optimizer and learning rate are adjusted as needed to ensure the stability and performance of the model.

During the fine-tuning of the BERT model, key formulas include the loss functions for the Masked Language Modeling and Next Sentence Prediction tasks. By minimizing this loss function, the model learns rich language representations, enabling it to achieve outstanding performance on downstream tasks. 


\label{subsec:Original Datasets}
\begin{table}[htbp]
\caption{Overview of raw dataset statistics}
\label{table:data_statistics_raw}
\centering
\small
\renewcommand{\arraystretch}{0.9}
\setlength{\tabcolsep}{8pt} 
\begin{tabularx}{0.95\linewidth}{@{}l >{\raggedright\arraybackslash}X@{}} 
\toprule
\textbf{Dataset} & \textbf{Source} \\
\midrule
\addlinespace[2pt]
\multicolumn{2}{@{}l}{\emph{SA}} \\ 
SST-2 & \huggingfacemap{stanfordnlp/sst2} \\
SST-5 & \huggingfacemap{SetFit/sst5} \\
IMDb & \huggingfacemap{stanfordnlp/imdb} \\
Yelp & \huggingfacemap{Yelp/yelp\_review\_full} \\
\midrule
\addlinespace[2pt]
\multicolumn{2}{@{}l}{\emph{Topic-CLS}} \\
AGNews & \huggingfacemap{fancyzhx/ag\_news} \\
20News & \huggingfacemap{SetFit/20\_newsgroups} \\
Yahoo & \huggingfacemap{community-datasets/yahoo\_answers\_topics} \\
\midrule
\addlinespace[2pt]
\multicolumn{2}{@{}l}{\emph{Q\&A}} \\
RaceQA & \huggingfacemap{ehovy/race} \\
MMLU & \huggingfacemap{cais/mmlu} \\
ARC & \huggingfacemap{allenai/ai2\_arc} \\
MMLU-Pro & \huggingfacemap{TIGER-Lab/MMLU-Pro} \\
\bottomrule
\end{tabularx}
\end{table}

\section{Experiments}
\label{sec:experiments}
Text classification tasks have significant practical value in the real world \cite{howard-ruder-2018-universal}. This study focuses on three typical text classification sub-tasks: SA, Topic-CLS, and Q\&A tasks. 

\begin{figure*}[t!]{}
\centering
\includegraphics[width=0.8\textwidth]{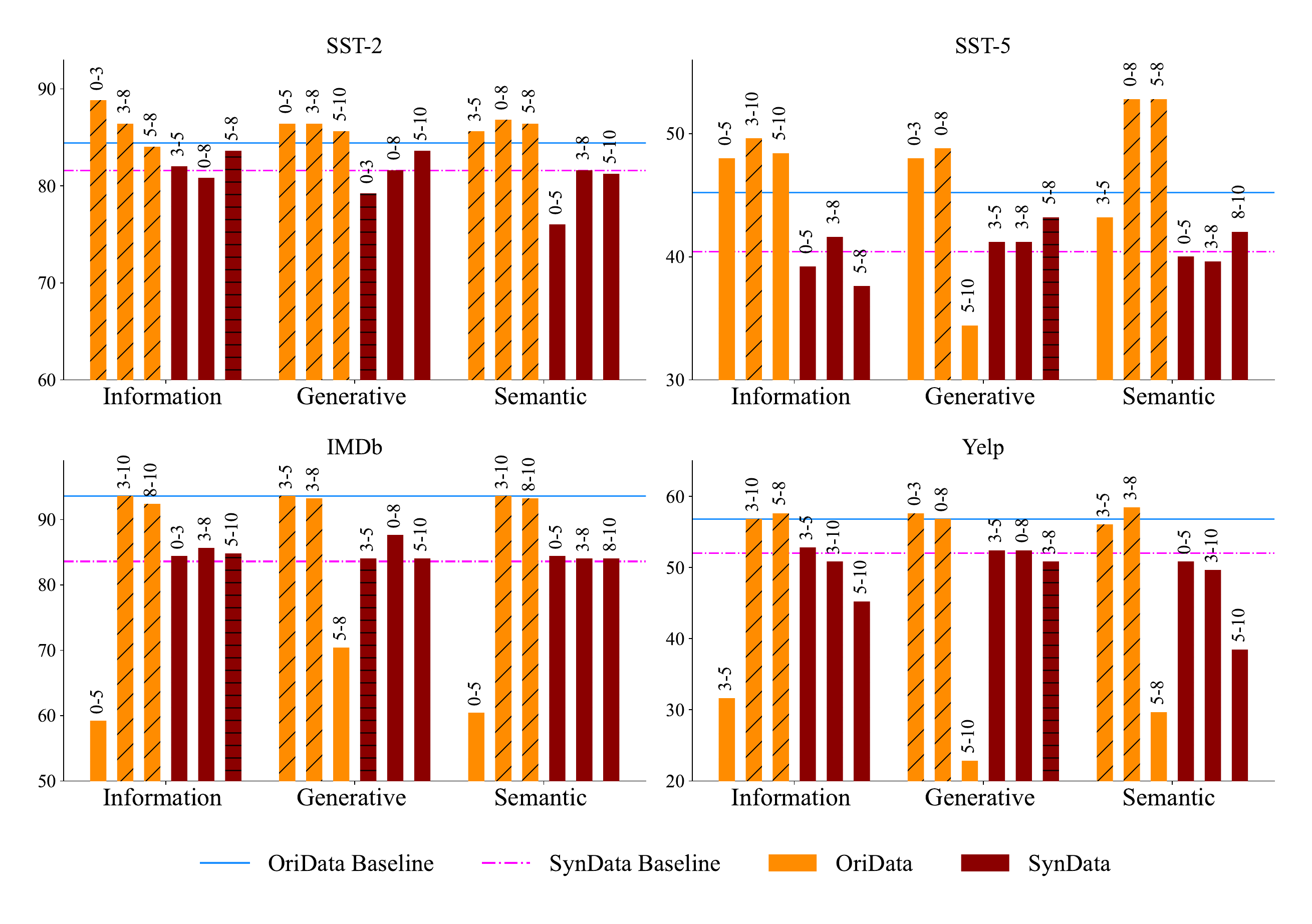}
\caption{The Effectiveness of EUDS in SA}
\label{fig:SA_data_dispaly}
\end{figure*}

\subsection{Original Datasets}
Our method is comprehensively evaluated using 11 widely studied datasets listed in Table \ref{table:data_statistics_raw}. The performance of the entropy-based data selection method is analyzed across different tasks, and performance differences between various entropy selection and baseline methods are compared. Particular focus is placed on the consumption of computational resources, data diversity, and improvements in model performance.

\textbf{SA}:\ For the SA task, our method is evaluated on the IMDb dataset \cite{maas-EtAl:2011:ACL-HLT2011}, the SST-2 and SST-5 datasets from the Stanford Sentiment Treebank \cite{socher-etal-2013-recursive}, and the Yelp dataset constructed by \cite{10.5555/2969239.2969312}.

\textbf{Topic-CLS}:\ For the Topic-CLS task, our method is evaluated on the 20News dataset \cite{Lang95}, the large-scale AGNews dataset \cite{10.5555/2969239.2969312}, and the Yahoo dataset \cite{10.5555/2969239.2969312}.

\textbf{Q\&A}:\ For the Q\&A task, we use large-scale reading comprehension datasets such as RaceQA \cite{lai-etal-2017-race}, the multiple-choice science question dataset ARC \cite{allenai:arc}, the large-scale multi-task language understanding dataset MMLU \cite{hendryckstest2021}, and the extended MMLU-based dataset MMLU-Pro \cite{wang2024mmlu}.

\subsection{Generating Synthetic Datasets}
\label{subsec:Synthetic Datasets}
GPT-4o is utilized to generate synthetic data (the dataset is recorded as SynData) for the 11 datasets involved in the three text classification sub-tasks described in Section \ref{subsec:Original Datasets}. To validate the effectiveness of the proposed EUDS method, the amounts of synthetic and original data are maintained consistently, with them being used separately as the training sets. Simultaneously, using the same validation and test sets, the models fine-tuned with synthetic and original data were evaluated.

\subsection{Technical Details}
\label{subsec:Technical Details}
The original data (the dataset is recorded as OriData) is extracted from the datasets described in Section \ref{subsec:Original Datasets}. In the single-entropy and combined-entropy validation experiments, 2,000 samples are extracted as OriData. In the single-entropy effectiveness verification experiment under the data expansion condition, 10,000 samples are extracted as OriData. For data partitioning, both OriData and SynData generated in Section \ref{subsec:Synthetic Datasets} are divided into training, validation, and test sets at an 8:1:1 ratio. The entropy value for each dataset is estimated using the EUDS framework.

This study focuses on models that can perform robustly across various tasks. To this end, we use the same hyperparameters for each task and adjust them based on the validation set. The BERT model is fine-tuned to perform single-label classification tasks. The training runs for 10 epochs, with a training batch size and evaluation batch size of 64 per device. The learning rate is set to 4e-5, the learning rate warm-up ratio is 0.1, and the weight decay is 0.1   \cite{DBLP:journals/corr/abs-1810-04805}.

\begin{figure*}[]{}
\centering
\includegraphics[width=0.8\textwidth]{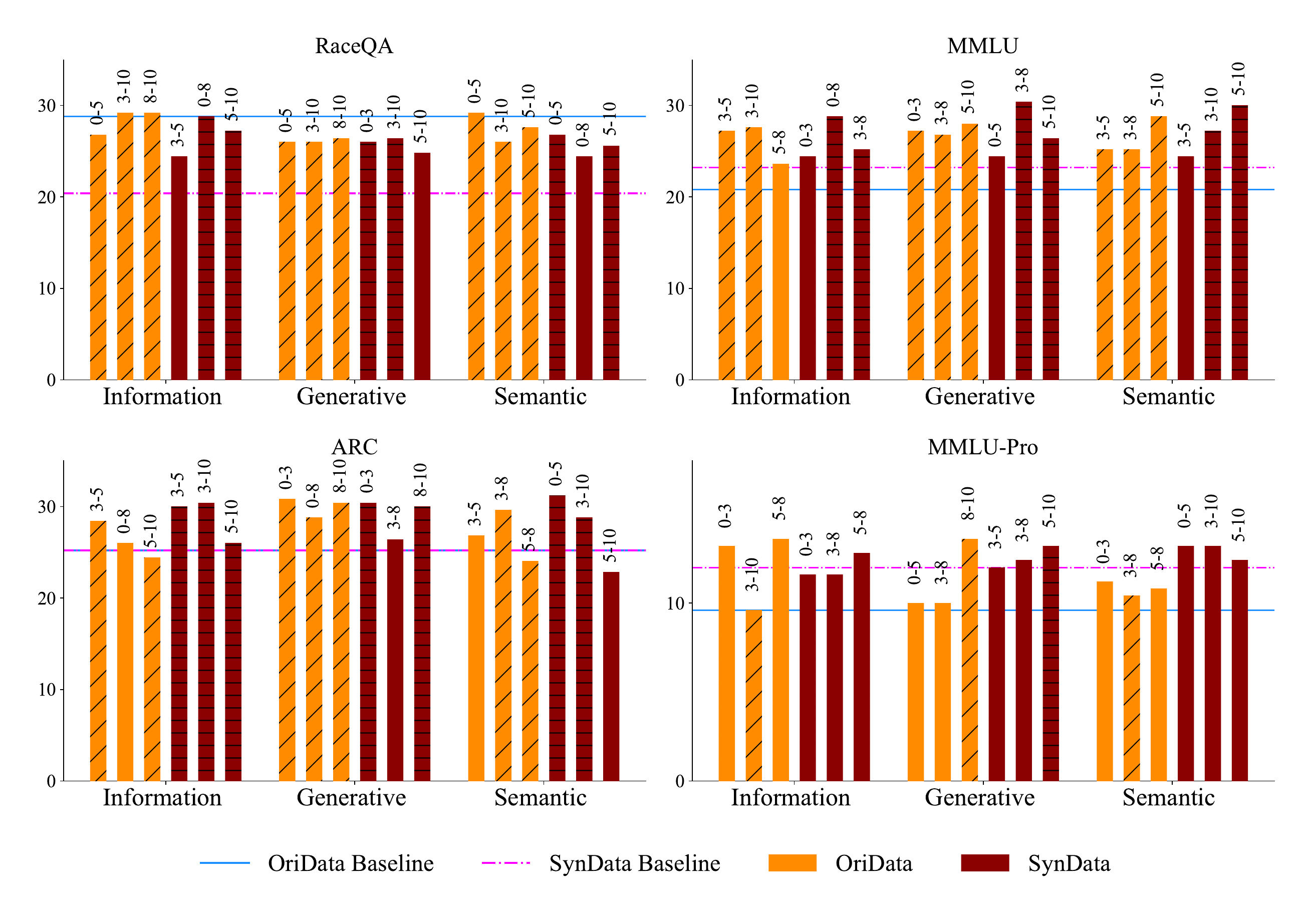}
\caption{The Effectiveness of EUDS in Q\&A}
\label{fig:QA_data_dispaly}
\end{figure*}

\begin{figure*}[]{}
\centering
\includegraphics[width=0.8\textwidth]{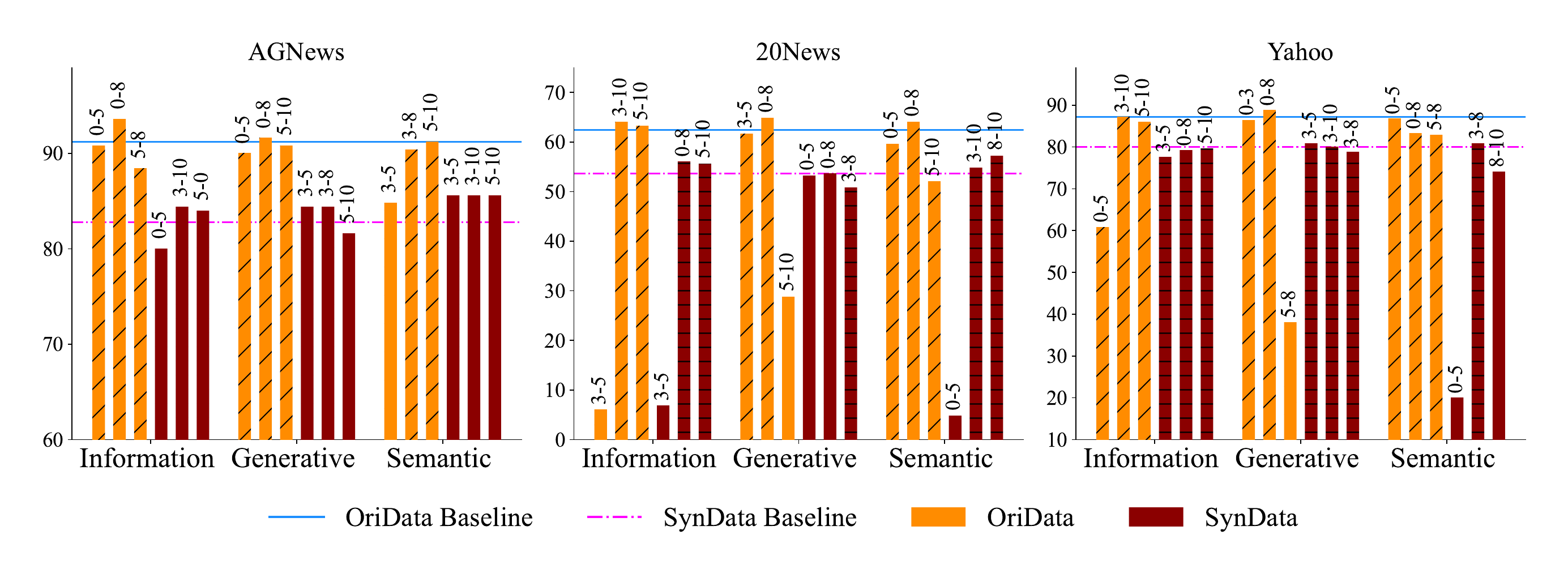}
\caption{The Effectiveness of EUDS in Topic-CLS}
\label{fig:TC_data_dispaly}
\end{figure*}

\subsection{Using a Single Entropy}
\label{subsec:Using a Single Entropy}
\subsubsection{Select OirData and SynData}
\label{subsub:Select OirData and SynData}
For the 11 datasets, data selection is performed using three methods: IE, GE, and SE. For each dataset, the entropy value of each sample is first estimated, and an effective data subset is selected based on a predefined threshold. Next, the selected data is used to fine-tune the BERT model, thereby validating the impact of different entropy selection methods on the model performance. The experimental results are presented in Figures \ref{fig:SA_data_dispaly}, \ref{fig:QA_data_dispaly} and \ref{fig:TC_data_dispaly}, clearly illustrating the performance comparison of different entropy selection methods across tasks.

\begin{table*}[]
\caption{The Best Interval Effect of Each Entropy After Selection of OriData.}
\begin{center}
\scalebox{0.85}{ 
\small 
\begin{threeparttable}
\begin{tabular}{lcccccccccccc}
\toprule
\textbf{OriData} & \multicolumn{4}{c}{\textbf{Information Entropy}} & \multicolumn{4}{c}{\textbf{Generative Entropy}} & \multicolumn{4}{c}{\textbf{Semantic Entropy}} \\ 
\cmidrule(lr){1-1} \cmidrule(lr){2-5} \cmidrule(lr){6-9} \cmidrule(lr){10-13}
\textbf{Tasks} & \textbf{Interval} & \textbf{Data} & \textbf{Accuracy} & \textbf{F1} & \textbf{Interval} & \textbf{Data} & \textbf{Accuracy} & \textbf{F1} & \textbf{Interval} & \textbf{Data} & \textbf{Accuracy} & \textbf{F1}\\ 
\midrule
SST-2 & 0-3 & 64.20↓ & 4.40↑ & 4.44↑ & 0-5 & 40.60↓ & 2.00↑ & 2.11↑ & 0-8 & 13.25↓ & 2.40↑ & 2.35↑  \\
SST-5 & 3-10 & 5.65↓ & 4.40↑ & 4.91↑ & 0-3 & 50.75↓ & 2.80↑ & 2.93↑ & 5-8 & 42.00↓ & 7.60↑ & 8.48↑ \\
IMDb & 3-10 & 0.20↓ & 0.00→ & 0.00→ & 3-5 & 33.55↓ & 0.00→ & 0.00→ & 3-10 & 0.50↓ & 0.00→ & 0.00→ \\
Yelp & 5-8 & 28.55↓ & 0.80↑ & 0.84↑ & 0-3 & 20.45↓ & 0.80↑ & 0.92↑ & 3-8 & 45.80↓ & 1.60↑ & 0.08↑ \\
AGNews & 0-8 & 0.70↓ & 2.40↑ & 2.36↑ & 0-8 & 0.25↓ & 0.40↑ & 0.38↑ & 5-10 & 13.50↓ & 0.00→ & 0.02↓ \\
20News & 3-10 & 1.65↓ & 1.60↑ & 1.79↑ & 0-8 & 0.65↓ & 2.40↑ & 2.43↑ & 0-8 & 0.75↓ & 1.60↑ & 1.65↑ \\
Yahoo & 3-10 & 0.90↓ & 0.00→ & 0.01↑ & 0-8 & 0.60↓ & 1.60↑ & 1.57↑ & 0-5 & 43.60↓ & 0.40↓ & 0.22↓ \\
RaceQA & 8-10 & 61.20↓ & 0.40↑ & 0.11↑ & 0-5 & 39.80↓ & 2.20↓ & 3.38↓ & 0-5 & 71.25↓ & 0.40↑ & 0.43↓  \\
MMLU & 3-10 & 48.10↓ & 6.60↑ & 5.49↑ & 5-10 & 40.00↓ & 7.20↑ & 7.37↑ & 5-10 & 52.75↓ & 8.00↑ & 8.46↑ \\
ARC & 3-5 & 57.65↓ & 3.20↑ & 0.71↑ & 0-3 & 78.70↓ & 5.60↑ & 9.57↓ & 3-8 & 64.75↓ & 4.40↑ & 1.58↑ \\
MMLU-Pro & 5-8 & 84.05↓ & 4.00↑ & 3.06↓ & 0-5 & 61.35↓ & 0.40↑ & 0.19↓ & 5-8 & 42.50↓ & 1.20↑ & 1.09↑ \\
\bottomrule
\end{tabular}
\begin{tablenotes}
    \small
    \item \textit{Note:} All values in the Data, Accuracy, and F1 columns are in percentages (\%). 
    The arrows indicate change direction: ↑ = increase, ↓ = decrease, → = no change.
\end{tablenotes}
\end{threeparttable}
}
\end{center}
\vspace{-1em}
\label{tab:main_oridata_display_table}
\end{table*}

\begin{table*}[]
\caption{The Best Interval Effect of Each Entropy After Selecting of SynData.}

\centering
\scalebox{0.85}
{
\small
\begin{threeparttable}
\begin{tabular}{lcccccccccccc}
\toprule
\textbf{SynData} & \multicolumn{4}{c}{\textbf{Information Entropy}} & \multicolumn{4}{c}{\textbf{Generative Entropy}} & \multicolumn{4}{c}{\textbf{Semantic Entropy}} \\
\cmidrule(lr){1-1} \cmidrule(lr){2-5} \cmidrule(lr){6-9} \cmidrule(lr){10-13}
\textbf{Tasks} & \textbf{Interval} & \textbf{Data} & \textbf{Accuracy} & \textbf{F1} & \textbf{Interval} & \textbf{Data} & \textbf{Accuracy} & \textbf{F1} & \textbf{Interval} & \textbf{Data} & \textbf{Accuracy} & \textbf{F1} \\
\midrule
SST-2 & 5-8 & 28.80↓ & 2.00↑ & 2.23↑ & 5-10 & 70.10↓ & 2.00↑ & 1.76↑ & 3-8 & 52.00↓ & 0.00→ & 0.25↑ \\
SST-5 & 3-8 & 11.15↓ & 1.20↑ & 0.50↑ & 5-8 & 78.80↓ & 2.80↑ & 0.48↓ & 8-10 & 58.50↓ & 1.60↑ & 2.18↑ \\
IMDb & 5-10 & 34.10↓ & 1.20↑ & 1.09↑ & 5-10 & 74.60↓ & 0.40↑ & 0.41↑ & 3-8 & 80.50↓ & 0.40↑ & 0.41↑ \\
Yelp & 3-5 & 63.40↓ & 0.80↑ & 0.81↓ & 3-5 & 54.65↓ & 0.40↑ & 0.48↓ & 0-5 & 17.70↓ & 1.20↓ & 1.79↓ \\
AGNews & 5-10 & 11.55↓ & 1.20↑ & 1.12↑ & 3-5 & 52.40↓ & 1.20↑ & 1.94↑ & 3-5 & 63.40↓ & 2.80↑ & 3.00↑ \\
20News & 0-8 & 15.30↓ & 2.40↑ & 2.14↑ & 0-8 & 0.30↓ & 0.40↑ & 0.08↓ & 8-10 & 2.95↓ & 2.60↑ & 3.87↑ \\
Yahoo & 5-10 & 23.70↓ & 0.40↓ & 0.23↓ & 3-5 & 69.00↓ & 0.80↑ & 0.72↑ & 3-8 & 62.05↓ & 0.80↑ & 1.02↑ \\
RaceQA & 3-5 & 13.45↓ & 4.00↑ & 1.12↓ & 0-3 & 67.80↓ & 5.60↑ & 1.94↑ & 0-8 & 66.35↓ & 4.00↑ & 4.04↓ \\
MMLU & 3-8 & 66.55↓ & 2.00↑ & 0.76↓ & 5-10 & 41.50↓ & 3.20↑ & 1.21↓ & 5-10 & 5.95↓ & 6.80↑ & 5.84↑ \\
ARC & 3-10 & 57.55↓ & 5.20↑ & 4.35↑ & 0-3 & 60.80↓ & 5.20↑ & 9.13↓ & 3-10 & 34.00↓ & 3.60↑ & 2.39↑ \\
MMLU-Pro & 5-10 & 89.40↓ & 0.80↑ & 4.06↓ & 5-10 & 80.15↓ & 1.20↑ & 2.22↓ & 0-5 & 80.70↓ & 1.20↑ & 1.93↓ \\
\bottomrule
\end{tabular}
\begin{tablenotes}
    \small
    \item \textit{Note:} All values in the Data, Accuracy, and F1 columns are in percentages (\%). 
    The arrows indicate change direction: ↑ = increase, ↓ = decrease, → = no change.
\end{tablenotes}
\end{threeparttable}
}

\vspace{-1em}
\label{tab:main_syndata_display_table}
\end{table*}

The experimental results of data selection, presented in Tables \ref{tab:main_oridata_display_table} and \ref{tab:main_syndata_display_table}, demonstrate that the optimal performance range (i.e., the balance between the data reduction rate and accuracy gain) is achieved by the three methods. The key characteristics are as follows:

\begin{itemize}
    \item[$\bullet$] \textbf{General Advantages} 
    \\
    \textbf{Efficient Data Reduction}: Using a single entropy method significantly reduces the training sample size while maintaining model performance. The experimental results show that in 198 optimal performance intervals across 22 datasets (both OriData and SynData), 65.66\% (130/198) of the scenarios achieved model performance that met or exceeded the baseline levels, and 15.15\% (30/198) of the scenarios maintained performance within a 1.5\% less error margin of the baseline. Typical examples include the following: in the MMLU-Pro task, IE selection reduced the data volume by 84.05\%; in the RaceQA task, SE selection reduced the data volume by 71.25\% (see Table \ref{table:data_statistics_raw}). 
    \\
    \textbf{Performance Optimization Effect}: The significant reduction in data volume does not lead to performance degradation; in some tasks, performance even improve. For example, in the MMLU-Pro task, the accuracy of OriData increased by 4.00\% after IE selection (with a data volume reduction of 84.05\%); in the RaceQA task, the accuracy increased by 0.40\% after SE selection (with a data volume reduction of 71.25\%). This result demonstrates that our method effectively identifies high-information training samples and breaks the ``data volume-performance" linear dependency. 
    \\
    \textbf{Applicability in Multiple Scenarios}: Even under non-resource-constrained conditions, this method still demonstrates optimization potential. For instance, in the AGNews task, IE selection reduce the data volume by only 0.70\%, but achieve a 2.40\% accuracy improvement. This indicates that the method offers the dual benefits of resource savings and performance enhancement. 
    \\
    \item[$\bullet$] \textbf{Limitations}
    \\
    \textbf {Marginal Performance Decline}: In some datasets, model performance show marginal decline after selection. For instance, in the Yahoo dataset, SynData is reduced by 23.70\% after IE selection, but the accuracy drop by 0.40\% compared to the baseline. This suggests that a balance between performance and efficiency must be achieved during data selection. 
    \\
    \textbf{Concentrated Entropy Distribution}: Some OriData and SynData exhibit concentrated entropy distribution characteristics. For example, in the MMLU-Pro SynData, the IE and GE selection samples are concentrated in the low-entropy range, whereas the SE selection samples tend toward the high-entropy range. This distribution characteristic limits the extent of data reduction; therefore, a joint evaluation of multidimensional entropy is required for better data selection.  
\end{itemize}

\subsubsection{Use of SynData and OriData Combinations}
\label{subsubsec:Use of SynData and OriData Combinations}
To address the challenge of limited data resources, this study adopts synthetic data generation as a solution. SynData usually needs to be used in conjunction with OriData. As described in Section \ref{subsub:Select OirData and SynData}, preliminary experiments only validated the independent applicability of the data selection method to the synthetic data. This Section further expands the experimental design to specifically evaluate the impact of the following two data selection combination strategies on model performance: (1) combining the selected SynData with the OriData (the dataset is recorded as SumData); (2) combining the selected SynData with the selected OriData (the dataset is recorded as JoSelData), to explore the synergistic effect of the dual selection strategy.

\begin{table*}[]
    \caption{Filtering the Effect Demonstration of the Optimal Interval for Generating SynData for an Established Task Gain.}
    \centering
    \small 
    \scalebox{0.85}{
    \begin{threeparttable}
    \begin{tabular}{lcccccccccccc}
        \toprule
        \multirow{2}{*}{\textbf{Tasks}}  & \multicolumn{4}{c}{\textbf{Information Entropy}} & \multicolumn{4}{c}{\textbf{Generative Entropy}} & \multicolumn{4}{c}{\textbf{Semantic Entropy}} \\
        \cmidrule(lr){2-5} \cmidrule(lr){6-9} \cmidrule(lr){10-13}
        & \textbf{Interval} & \textbf{Data} & \textbf{Accuracy} & \textbf{F1} & \textbf{Interval} & \textbf{Data} & \textbf{Accuracy} & \textbf{F1} & \textbf{Interval} & \textbf{Data} & \textbf{Accuracy} & \textbf{F1}\\
        \midrule
        SST-2 & 0-3 & 46.82↓ & 0.00→ & 0.03↓ & 5-8 & 35.85↓ & 0.80↓ & 0.90↓ & 3-5 & 46.25↓ & 0.00→ & 0.05↓ \\
        SST-5 & 3-8 & 5.57↓ & 0.80↓ & 0.02↑ & 5-8 & 39.40↓ & 0.40↓ & 0.25↓ & 5-8 & 26.25↓ & 2.00↑ & 1.69↑ \\
        IMDb & 3-5 & 35.20↓ & 3.20↑ & 3.21↑ & 5-8 & 37.85↓ & 2.80↑ & 2.81↑ & 5-8 & 43.25↓ & 3.20↑ & 3.21↑ \\
        Yelp & 5-8 & 24.10↓ & 1.20↓ & 1.07↓ & 3-10 & 19.17↓ & 2.40↓ & 2.41↓ & 3-5 & 9.50↓ & 1.60↓ & 1.78↓ \\
        AGNews & 0-8 & 5.47↓ & 0.40↑ & 0.41↑ & 0-5 & 6.62↓ & 0.80↑ & 0.81↑ & 5-10 & 18.52↓ & 0.40↓ & 0.43↓ \\
        20News & 3-8 & 7.67↓ & 1.60↑ & 2.04↑ & 3-10 & 22.30↓ & 1.20↑ & 1.19↑ & 5-8 & 48.57↓ & 1.20↑ & 1.31↑ \\
        Yahoo & 0-3 & 48.52↓ & 1.60↑ & 1.51↑ & 3-10 & 32.70↓ & 1.20↑ & 1.47↑ & 3-5 & 49.95↓ & 2.40↑ & 2.65↑ \\
        RaceQA & 3-8 & 5.10↓ & 2.80↑ & 2.60↑ & 0-5 & 5.90↓ & 3.20↑ & 4.20↑ & 8-10 & 17.97↓ & 6.80↑ & 5.90↑ \\
        MMLU & 8-10 & 49.67↓ & 4.80↑ & 3.52↑ & 3-10 & 0.72↓ & 5.20↑ & 4.91↑ & 8-10 & 15.22↓ & 5.60↑ & 5.33↑ \\
        ARC & 8-10 & 49.47↓ & 0.00→ & 0.49↓ & 8-10 & 48.07↓ & 0.80↓ & 5.42↓ & 3-5 & 49.70↓ & 0.80↑ & 0.11↑ \\
        MMLU-Pro & 3-5 & 38.52↓ & 3.60↑ & 1.01↑ & 5-8 & 40.10↓ & 4.40↑ & 3.00↑ & 5-8 & 30.17↓ & 0.40↑ & 1.17↓ \\
        \bottomrule
    \end{tabular}
    \begin{tablenotes}
    \small
    \item \textit{Note:} All values in the Data, Accuracy, and F1 columns are in percentages (\%). 
    The arrows indicate change direction: ↑ = increase, ↓ = decrease, → = no change.
\end{tablenotes}
\end{threeparttable}
    }

    \label{tab:synthetic_select_performance}
\end{table*}

For combined strategy (1), we systematically evaluate the effectiveness of the data selection method in enhancing the performance of predefined classification tasks using selective Syndata. The experimental results, as shown in Table \ref{tab:synthetic_select_performance}, demonstrate that our method exhibits significant advantages across various text classification sub-tasks.

Despite significantly reducing the amount of data, our method can maintain or even surpass the baseline model’s performance.  
For example, in the MMLU dataset, after IE selection, the data volume is reduced by 49.67\%, and the accuracy improved by 4.80\%; in the MMLU-Pro dataset, after GE selection, the data volume is reduced by 40.10\%, and the accuracy improve by 4.40\%; in the ARC dataset, after SE selection, the data volume is reduced by 49.70\%, and the accuracy improved by 0.80\%.  These results demonstrate that our entropy-based selection mechanism effectively identifies valuable samples in synthetic data, breaking the traditional data volume-model performance linear dependency. 

Although in some cases, the selection strategy does not significantly enhance the model performance, we successfully reduce the computational resource consumption by significantly reducing the data volume. This suggests that in specific task scenarios, the selection strategy can act as a regulatory lever between ``computational resources-model performance," providing a practical solution for resource-constrained environments. The EUDS method not only significantly reduces the data volume but also enhances or maintains model performance, offering effective support for resource-constrained environments and showing clear benefits in predefined classification tasks.

To further reduce the amount of data used based on combined strategy (1), combined strategy (2) is adopted where the selected OriData were merged with SynData from the same range. The impact of the data selection strategy on the model performance is then evaluated alongside the reduction in the data scale.
The experimental results, as shown in Table \ref{tab:base_synthetic_select_performance}, detail the optimal performance of the different tasks under different data selection strategies. A comparative analysis with Table \ref{tab:synthetic_select_performance} indicates that for some tasks, the data volume can be significantly reduce while still maintaining or surpassing the baseline performance. Owing to inconsistencies in the selected samples, there is some deviation in the matching of effective intervals the same time. This study only explored the effect of merging data from the same range, and the combined effect of data from different ranges requires further investigation.

\begin{table*}[]
\caption{Display of best results interval after screening OriData and SynData.}
\begin{center}
\centering
\scalebox{0.85}
{
\small
\begin{threeparttable}
\begin{tabular}{lcccccccccccc}
\toprule
\multirow{2}{*}{\textbf{Tasks}} & \multicolumn{4}{c}{\textbf{Information Entropy}} & \multicolumn{4}{c}{\textbf{Generative Entropy}} & \multicolumn{4}{c}{\textbf{Semantic Entropy}} \\ 
\cmidrule(lr){2-5} \cmidrule(lr){6-9} \cmidrule(lr){10-13}
& \textbf{Interval} & \textbf{Data} & \textbf{Accuracy} & \textbf{F1} & \textbf{Interval} & \textbf{Data} & \textbf{Accuracy} & \textbf{F1} & \textbf{Interval} & \textbf{Data} & \textbf{Accuracy} & \textbf{F1} \\ 
\midrule
SST-2 & 0-5 & 57.92↓ & 0.40↑ & 0.73↑ & 3-10 & 25.00↓ & 1.60↓ & 1.54↓ & 5-8 & 54.12↓ & 0.80↓ & 0.84↓ \\
SST-5 & 3-10 & 6.60↓ & 1.20↓ & 1.13↓ & 0-5 & 16.00↓ & 1.20↓ & 0.55↓ & 3-8 & 37.25↓ & 0.80↑ & 1.94↑ \\
IMDb & 8-10 & 60.27↓ & 2.00↑ & 2.01↑ & 3-8 & 25.77↓ & 3.20↑ & 3.21↑ & 8-10 & 30.12↓ & 2.40↑ & 2.41↑ \\
Yelp & 3-8 & 15.05↓ & 2.40↓ & 2.75↓ & 3-8 & 59.80↓ & 1.20↓ & 0.99↓ & 0-5 & 11.22↓ & 0.80↑ & 0.79↑ \\
AGNews & 3-8 & 11.97↓ & 0.40↓ & 0.32↓ & 0-5 & 17.30↓ & 1.20↓ & 1.30↓ & 5-8 & 65.45↓ & 1.60↓ & 1.71↓ \\
20News & 5-8 & 33.50↓ & 0.00→ & 0.15↑ & 0-8 & 0.47↓ & 0.00→ & 0.43↑ & 0-5 & 68.20↓ & 0.40↓ & 0.87↓ \\
Yahoo & 3-10 & 1.95↓ & 0.80↑ & 1.00↑ & 3-10 & 62.80↓ & 0.40↑ & 0.51↑ & 0-5 & 71.67↓ & 0.40↓ & 0.18↓ \\
RaceQA & 5-8 & 73.47↓ & 3.20↑ & 3.24↑ & 3-5 & 42.40↓ & 2.00↑ & 1.47↑ & 3-8 & 40.55↓ & 2.20↑ & 3.25↑ \\
MMLU & 3-10 & 57.00↓ & 8.80↑ & 8.85↑ & 0-5 & 59.25↓ & 4.40↑ & 4.44↑ & 8-10 & 64.72↓ & 6.80↑ & 4.45↑ \\
ARC & 0-5 & 24.40↓ & 0.80↑ & 0.62↑ & 0-3 & 94.35↓ & 2.40↑ & 15.01↓ & 0-8 & 73.20↓ & 1.20↑ & 0.21↑ \\
MMLU-Pro & 3-10 & 63.00↓ & 6.00↑ & 1.76↓ & 5-10 & 59.45↓ & 4.40↑ & 2.08↑ & 5-10 & 34.27↓ & 3.20↑ & 0.36↓ \\
\bottomrule
\end{tabular}
    \begin{tablenotes}
    \small
    \item \textit{Note:} All values in the Data, Accuracy, and F1 columns are in percentages (\%). 
    The arrows indicate change direction: ↑ = increase, ↓ = decrease, → = no change.
\end{tablenotes}
\end{threeparttable}
}
\end{center}
\vspace{-1em}
\label{tab:base_synthetic_select_performance}
\end{table*}

\subsubsection{Data Augmentation}
Generating synthetic data effectively alleviates the problem of limited data resources, and expanding the scale of language model (LM) fine-tuning data further enhances generalization performance across diverse language tasks \cite{sanh2022multitaskpromptedtrainingenables, JMLR:v25:23-0870, 10.5555/3618408.3619349}. Computational resource limitations are a significant challenge. Efficiently performing data selection under large-scale data conditions is an urgent issue that must be addressed. As previously mentioned, the effectiveness of the EUDS method is validated using a small-scale dataset (2,000 data points). To further assess its applicability to large-scale datasets, experiments were conducted on an expanded dataset (10,000 data points, five times larger) to evaluate the robustness and effectiveness of the data-selection method under different data scales.

\begin{table*}[]
\caption{Demonstration of the effects of applying our data selection method at a larger data scale. Table showing changes in accuracy and volume of data.}
\begin{center}
\centering
\resizebox{\linewidth}{!}{
\small
\begin{threeparttable}
\begin{tabular}{lccccccccccccc}
\toprule 
\multirow{2}{*}{\textbf{Tasks}} & \multicolumn{4}{c}{\textbf{Information Entropy}} & \multicolumn{4}{c}{\textbf{Generative Entropy}} & \multicolumn{4}{c}{\textbf{Semantic Entropy}} \\
\cmidrule(lr){2-5} \cmidrule(lr){6-9} \cmidrule(lr){10-13}
& \textbf{OriData} & \textbf{SynData} & \textbf{SumData} & \textbf{JoSelData} & \textbf{OriData} & \textbf{SynData} & \textbf{SumData} & \textbf{JoSelData} & \textbf{OriData} & \textbf{SynData} & \textbf{SumData} & \textbf{JoSelData}\\
\midrule
SST-5 & 25.00↓\textsubscript{1.28↑} & 45.00↓\textsubscript{1.92↑}& 48.00↓\textsubscript{2.40↑} & 10.22↓\textsubscript{0.08↑} & 10.88↓\textsubscript{0.32↓} & 74.11↓\textsubscript{2.96↑} & 49.91↓\textsubscript{2.56↑} & 65.36↓\textsubscript{0.48↑} & 29.40↓\textsubscript{0.08↑} & 87.18↓\textsubscript{0.08↑} & 49.99↓\textsubscript{2.24↑} & 58.30↓\textsubscript{1.52↑}\\
AGNews & 52.83↓\textsubscript{0.00↑} & 76.03↓\textsubscript{2.88↑}& 38.01↓\textsubscript{1.72↑} & 87.82↓\textsubscript{8.08↑} & 20.65↓\textsubscript{0.24↓} & 41.74↓\textsubscript{1.60↑} & 49.96↓\textsubscript{5.12↑} & 60.61↓\textsubscript{5.04↑} & 65.10↓\textsubscript{0.40↑} & 0.06↓\textsubscript{1.12↑}  & 
0.28↓\textsubscript{2.24↑}  & 18.03↓\textsubscript{1.84↑}\\
MMLU & 51.51↓\textsubscript{3.84↑} & 85.71↓\textsubscript{0.64↑} & 42.85↓\textsubscript{1.68↑} & 89.38↓\textsubscript{1.68↑} & 70.00↓\textsubscript{0.08↑} & 90.87↓\textsubscript{0.56↑} &
45.43↓\textsubscript{0.16↑} & 62.17↓\textsubscript{0.12↑} & 40.85↓\textsubscript{1.44↑} & 87.27↓\textsubscript{0.80↑} & 49.51↓\textsubscript{1.94↑} & 72.25↓\textsubscript{0.88↑}\\
\bottomrule
\end{tabular}
 \begin{tablenotes}
    \small
    \item \textit{Note:} The first value in each cell represents accuracy change in percentage points; the second value represents data reduction percentage. Arrows indicate direction: ↑ = increase, ↓ = decrease, → = no change.
\end{tablenotes}
\end{threeparttable}
}
\end{center}
\vspace{-1em}
\label{tab:data_augmentation}
\end{table*}

As shown in Table \ref{tab:data_augmentation}, the method still maintains excellent selection efficiency on large-scale datasets, demonstrating its generalization ability across different data scales. The issue of data distribution is also noteworthy. Specifically, some large-scale SynData exhibit a more pronounced feature concentration phenomenon than small-scale data after entropy-based selection. For example, in the AGNews dataset, after SE selection, 99.44\% of the SynData is concentrated in the high-entropy range of 8-10, and when combined with the OriData, this proportion increases further to 99.72\%. Compared with small-scale data, this phenomenon may limit the effectiveness of data selection.

\subsection{Using Combined Entropy}
\label{subsec:Using Combined Entropy}
The combination entropy method is adopted for a more refined data selection to minimize the consumption of computing and data resources. Specifically, various combination schemes are designed, including combinations of IE and GE (IGE), IE and SE (ISE), GE and SE (GSE), and the joint application of all three entropies: IE, GE, and SE (IGSE). These different combination entropy methods are utilized to fine-tune the BERT model, and explored their effects on alleviating computational resource limitations.

As shown in Section \ref{subsub:Select OirData and SynData}, the single entropy method has demonstrated effectiveness in specific classification tasks. Owing to limitations in computational resources, we developed a more efficient combined entropy selection strategy. This strategy accurately quantifies the data value density by jointly evaluating multiple entropy values.

\begin{table*}[]
\caption{Application of Optimal Interval Effect of Combined Entropy to OriData and SynData Respectively. Table showing changes in accuracy and volume of data.}
\centering
\scalebox{0.95}{
\small
\begin{threeparttable}
\begin{tabular}{lcccccccccc}
\toprule
\multirow{2}{*}{\textbf{Tasks}} & \multicolumn{2}{c}{\textbf{IG Entropy}} & \multicolumn{2}{c}{\textbf{IS Entropy}} & \multicolumn{2}{c}{\textbf{GS Entropy}} & \multicolumn{2}{c}{\textbf{IGS Entropy}} \\
\cmidrule(lr){2-3} \cmidrule(lr){4-5} \cmidrule(lr){6-7} \cmidrule(lr){8-9}
& \textbf{OriData} & \textbf{SynData} & \textbf{OriData} & \textbf{SynData} & \textbf{OriData} & \textbf{SynData} & \textbf{OriData} & \textbf{SynData} \\
\midrule
SST-5 & 22.65↓\textsubscript{0.40↑} & 29.60↓\textsubscript{2.40↑} & 32.20↓\textsubscript{4.80↑} & 53.35↓\textsubscript{0.80↑} & 64.75↓\textsubscript{1.60↑} & 24.65↓\textsubscript{0.40↑} & 45.00↓\textsubscript{2.40↑} & 31.90↓\textsubscript{0.40↑} \\
AGNews & 55.00↓\textsubscript{0.00↑} & 44.35↓\textsubscript{0.80↑} & 14.10↓\textsubscript{0.40↓} & 38.35↓\textsubscript{1.60↑} & 25.35↓\textsubscript{0.00↑} & 58.70↓\textsubscript{1.20↑} & 52.30↓\textsubscript{0.80↓} & 62.30↓\textsubscript{2.00↑} \\
MMLU & 55.35↓\textsubscript{2.00↑} & 66.70↓\textsubscript{0.00↑} & 60.00↓\textsubscript{8.00↑} & 67.50↓\textsubscript{0.80↑} & 72.35↓\textsubscript{4.80↑} & 70.35↓\textsubscript{2.40↑} & 61.25↓\textsubscript{4.80↑} & 70.95↓\textsubscript{4.50↑} \\
\midrule
& \textbf{SumData} & \textbf{JoSelData} & \textbf{SumData} & \textbf{JoSelData} & \textbf{SumData} & \textbf{JoSelData} & \textbf{SumData} & \textbf{JoSelData} \\
\midrule
SST-5 & 50.00↓\textsubscript{0.40↓} & 13.62↓\textsubscript{2.80↓} & 38.40↓\textsubscript{0.80↓} & 44.62↓\textsubscript{0.80↓} & 44.97↓\textsubscript{0.80↑} & 34.52↓\textsubscript{4.00↑} & 50.00↓\textsubscript{0.40↓} & 44.95↓\textsubscript{0.80↓} \\
AGNews & 47.17↓\textsubscript{1.20↓} & 74.67↓\textsubscript{0.80↓} & 48.92↓\textsubscript{1.60↓} & 45.25↓\textsubscript{0.40↓} & 48.45↓\textsubscript{0.00↓} & 40.87↓\textsubscript{2.00↓} & 47.70↓\textsubscript{0.40↓} & 45.60↓\textsubscript{2.00↓} \\
MMLU & 33.87↓\textsubscript{3.60↑} & 61.55↓\textsubscript{6.40↑} & 46.32↓\textsubscript{6.80↑} & 61.77↓\textsubscript{5.20↑} & 36.40↓\textsubscript{5.20↑} & 79.17↓\textsubscript{8.40↑} & 44.40↓\textsubscript{6.80↑} & 38.20↓\textsubscript{8.80↑} \\
\bottomrule
\end{tabular}
\begin{tablenotes}
    \small
    \item \textit{Note:} The first value in each cell represents accuracy change in percentage points; the second value represents data reduction percentage. Arrows indicate direction: ↑ = increase, ↓ = decrease, → = no change.
\end{tablenotes}
\end{threeparttable}
}
\vspace{-1em}
\label{tab:combined_entropy}
\end{table*}

The experimental results, as shown in Table \ref{tab:combined_entropy} and the comparative analysis of Tables \ref{tab:main_oridata_display_table} and \ref{tab:main_syndata_display_table}, indicate that in the optimal performance range, the combined entropy method shows significant advantages over the single entropy method. Through more precise selection, the number of samples required for model training is significantly reduced, thereby lowering data storage and processing costs and improving training efficiency. Since we focused only on the optimal performance range, there are still a few cases where the combined entropy method did not surpass the performance of the single entropy method.

\subsection{Comprehensive Promotion of Subsets}

\begin{table*}[htbp]
\centering
\small
\caption{The Impact of Applying the Same Entropy Interval to Both Full Dataset and Subset on Model Performance and Data Reduction.}
\label{tab:entropy_best_intervals}
\renewcommand{\arraystretch}{1.3}
\setlength{\tabcolsep}{8pt}
\begin{tabular}{llcccccc}
\toprule
\textbf{Entropy} & \textbf{Dataset} & \textbf{Interval} & \makecell[c]{Subset \\ Acc ↑ (\%)} & \makecell[c]{Subset \\ Reduce (\%)} & \makecell[c]{Fullset \\ Acc ↑ (\%)} & \makecell[c]{Fullset \\ Reduce (\%)} \\
\midrule
\multirow{4}{*}{IE} 
       & OriData   & 0-3 & 7.69↑ & 51.85↓ & 7.49↑ & 66.56↓ \\
       & SynData   & 0-3 & 5.17↑ & 34.10↓ & 3.74↑ & 16.97↓ \\
       & SumData   & 5-8 & 20.75↑ & 45.63↓ & 3.19↑ & 49.59↓ \\
       & JoSeData  & 0-3 & 16.98↑ & 42.97↓ & 1.60↑ & 41.76↓ \\
\midrule
\multirow{4}{*}{GE} 
       & OriData   & 0-5 & 21.15↑ & 60.00↓ & 0.33↑ & 69.06↓ \\
       & SynData   & 5-8 & 5.17↑ & 42.95↓ & 0.31↑ & 90.97↓ \\
       & SumData   & 3-5 & 13.21↑ & 29.97↓ & 5.11↑ & 20.49↓ \\
       & JoSeData  & 0-3 & 28.30↑ & 94.88↓ & 2.24↑ & 84.50↓ \\
\midrule
\multirow{4}{*}{SE} 
       & OriData   & 3-8 & 21.15↑ & 16.50↓ & 5.54↑ & 27.45↓ \\
       & SynData   & 8-10 & 10.34↑ & 30.45↓ & 3.12↑ & 14.59↓ \\
       & SumData   & 3-5 & 24.53↑ & 48.08↓ & 0.96↑ & 49.18↓ \\
       & JoSeData  & 0-8 & 16.98↑ & 35.27↓ & 3.83↑ & 50.51↓ \\
\bottomrule
\end{tabular}
\end{table*}

As previously discussed, we have verified that EUDS is effective on both small-scale and large-scale datasets. In our experiments, the smaller datasets are deliberately selected subsets of the corresponding large datasets—that is, the large datasets serve as the full sets, and the small datasets are their subsets. As shown in Table \ref{tab:entropy_best_intervals}, the optimal entropy intervals identified on the subsets often align with those of the full sets, both contributing to improved model performance. This observation suggests that, when determining the optimal entropy interval for a large-scale dataset, it is possible to first identify it on a representative subset and then generalize the result to the entire dataset. Therefore, to determine the optimal entropy interval for data selection in large-scale datasets, we only need to identify it on a representative subset and apply this interval to the full dataset for fine-tuning. This greatly reduces computational cost while preserving selection quality.

\section{Discussion}

\begin{table}[htbp]
\centering
\small
\caption{Comparison of EUDS with recent SOTA methods on representative datasets.}
\label{table:sota_dataset_comparison}
\renewcommand{\arraystretch}{1.25}
\setlength{\tabcolsep}{2pt} 
\begin{tabularx}{0.90\columnwidth}{
  >{\raggedright\arraybackslash}X
  >{\raggedright\arraybackslash}b{3.5cm} 
  >{\centering\arraybackslash}m{1.8cm} 
}
\toprule
Dataset & Methods & Accuracy (\%) \\
\midrule
\multirow{7}{*}{IMDb} 
   & BiLSTM & 81.87 \\
   & RoBERTa-BiLSTM & 92.46 \\
   & TWSSenti & 95.00 \\
   & SELECT & 63.79 \\
   & \textbf{IE (Ours)} & \textbf{93.60} \\
   & \textbf{GE (Ours)} & \textbf{93.20} \\
   & \textbf{SE (Ours)} & \textbf{93.60} \\
\cmidrule(lr){1-3}
\multirow{5}{*}{AGNews}
   & SELECT & 80.05 \\
   & FDKT & 75.60 \\
   & \textbf{IE (Ours)} & \textbf{93.60} \\
   & \textbf{GE (Ours)} & \textbf{91.60} \\
   & \textbf{SE (Ours)} & \textbf{91.20} \\
\bottomrule
\end{tabularx}
\end{table}

We compared the proposed EUDS method with several mainstream approaches, including BiLSTM \cite{nkhata2025finetuningbertbidirectionallstm}, RoBERTa-BiLSTM \cite{Rahman_2025} , TWSSenti \cite{albladi2025twssentinovelhybridframework}, SELECT \cite{liu2024tsds}, and FDKT\cite{li2024federateddomainspecificknowledgetransfer}. As shown in Table \ref{table:sota_dataset_comparison}, EUDS consistently outperforms existing methods in most cases. Even in the rare cases where the performance gap is minimal, EUDS remains highly competitive. These results strongly demonstrate the effectiveness and robustness of our approach across diverse data scenarios.

Taking IE as an example, the fundamental mechanism underlying the effectiveness of our entropy-based sample selection can be elucidated. Specifically, the $\mathrm{IE(x)}$ defined in Eq.~(2) measures the expected uncertainty or information content of a sample. Selecting samples within an optimal entropy interval thus balances diversity and stability, ensuring that the training data provides maximal informative content.
 
Moreover, the generalization error of a model can be bounded by a term involving the mutual information $\mathrm{I(D; \theta)}$ between the selected dataset $D$ and the model parameters $\theta$. By selecting data with balanced entropy, we effectively increase $\mathrm{I(D; \theta)}$, ensuring that each gradient update carries more informative signals, which accelerates convergence and enhances generalization. Formally, if $\mathcal{I^*}$ denotes the empirically determined optimal entropy interval, training on
\begin{equation}
\mathrm{D}_{\mathcal{I}^*} = \{x \in D \mid \mathrm{IE}(x) \in \mathcal{I}^*\}
\end{equation}
maximizes the expected marginal loss reduction:
\begin{equation}
\mathbb{E}_{x \sim \mathrm{D}_{\mathcal{I}^*}} \big[ \Delta \mathcal{L}(x) \big],
\end{equation}
where $\Delta \mathcal{L}(x)$ represents the contribution of sample $x$ to reducing training loss. This theoretical explanation aligns with our empirical findings that entropy-guided selection not only reduces computational cost but also improves or maintains model performance in large-scale fine-tuning.  

In the empirical exploration of this study, we thoroughly verify the effectiveness of the proposed entropy-based data selection method in text classification tasks through a series of rigorous experiments, providing strong support for its practical application. However, it must be noted that there are still certain limitations at the theoretical analysis level, which restrict our in-depth understanding of the essence of the method.

When facing different text classification sub-tasks, single or combined entropy is attempted for data selection. Unfortunately, a unified optimal entropy range suitable for all tasks cannot be found. This indicates that considerable room for optimization exists in the selection and application of entropy, necessitating Further theoretical exploration and explanation are required.

During the data selection process, whether for OriData or SynData, the use of a certain entropy for filtering may lead to a concentration in the data distribution. The consequences of this phenomenon are complex and twofold: on the one hand, although the model performance can be maintained or even improved to some extent, the reduction in data volume may not reach the ideal state, making it difficult to fully leverage the advantages of data selection in resource-constrained scenarios.  However, if excessive emphasis is placed on significantly reducing the data volume, the model may fail to converge. In this case, although the model performance seems to improve significantly, its stability and reliability are severely compromised, failing to meet the strict demands for model trustworthiness in practical applications.

Although our method achieves promising results at the experimental level, the lack of theoretical analysis and potential issues in the data selection process still require attention and further investigation in future research. We aim to continuously improve and optimize this data selection method to enable it to play a more outstanding and stable role in the field of text classification.

\section{Conclusions}
\label{sec:conclusions}
Data selection is a valuable tool for fine-tuning LMs, with the primary advantage of saving computational resource. In this study, we propose the EUDS. Although this method is relatively simple, it has demonstrated excellent performance in practical applications. Through fine-tuning experiments on the BERT model for three key sub-tasks in text classification, the results strongly indicate that, in the vast majority of application scenarios, our method can significantly reduce the amount of data required for fine-tuning the model, effectively alleviating resource constraints. 
Furthermore, the model's performance not only remains unchanged but, in some cases, it even surpasses the performance of the original model. All of this is achieved without sacrificing performance but rather through more efficient data utilization.  

Despite the positive results, there are still certain limitations that require further improvement and refinement. Nevertheless, we have provided a simple yet promising new method for the data selection field, which we hope will inspire further research into data selection methods that enhance the computational efficiency.

\bibliographystyle{IEEEtran}
\bibliography{references} 
\vspace{150pt}
\begin{IEEEbiography}[{\includegraphics[width=1in,height=1.25in,clip,keepaspectratio]{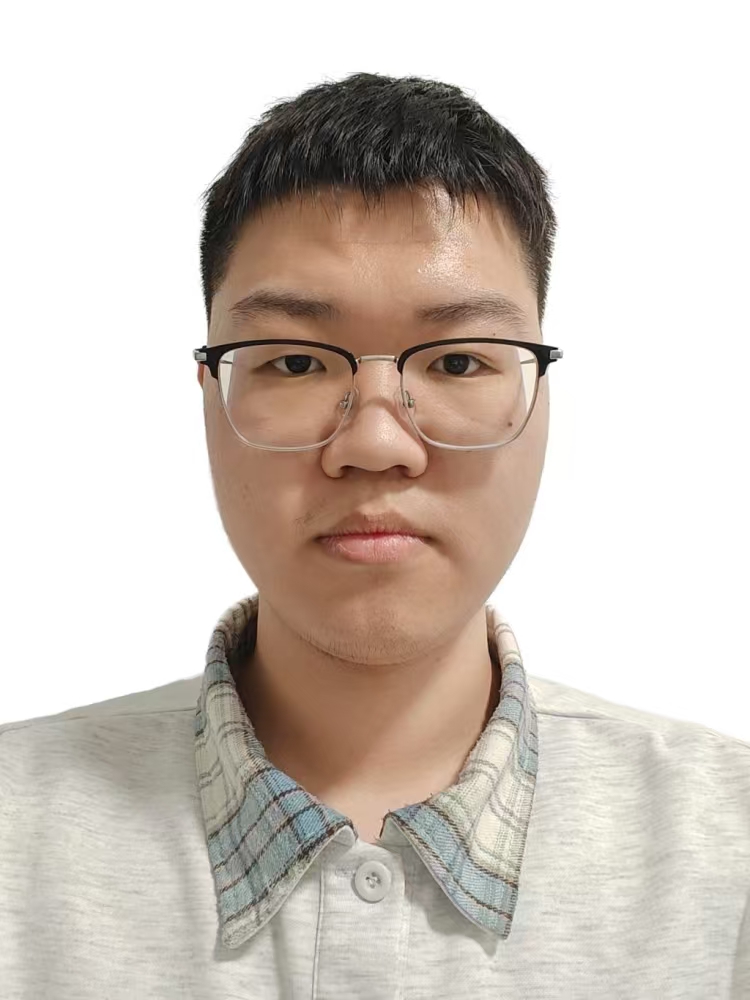}}]{Hongming Li} was born in Cangzhou, Hebei, China in 2003. He is currently studying in the School of Computer and Communication Engineering at University of Science and Technology Beijing, majoring in computer science and technology. In terms of work and study experience, he focuses on his study in the university currently. His research interests currently include natural language processing and computer vision.
\end{IEEEbiography}

\vspace{-250pt}
\begin{IEEEbiography}[{\includegraphics[width=1in,height=1.25in,clip,keepaspectratio]{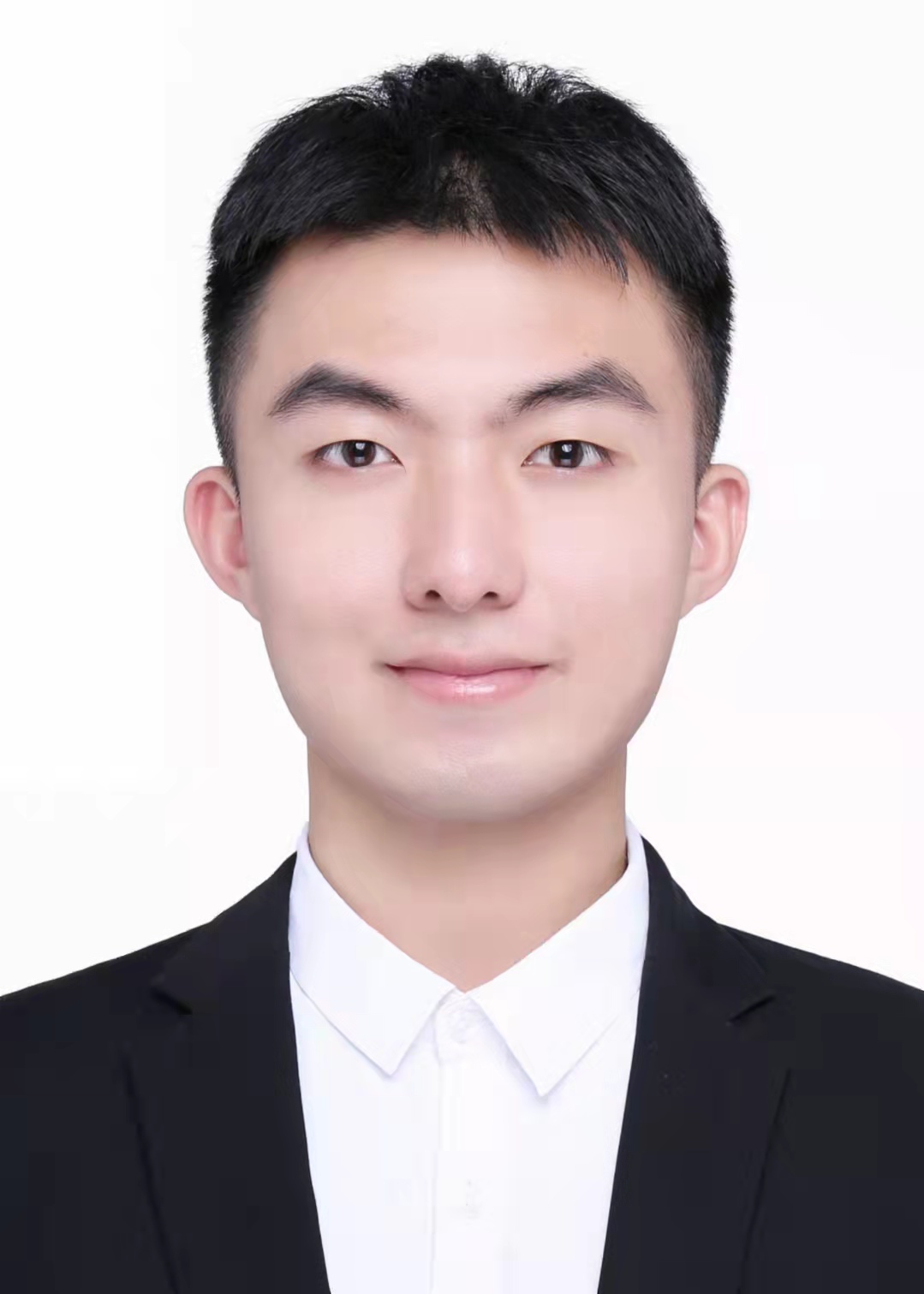}}]{Yang Liu} received his B.Eng. degree in Mechanical Engineering from University of Science and Technology Beijing, China, in 2021, M.S. degree in School of Computer and Communication Engineering from University of Science and Technology Beijing, China, in 2024.
He is currently a full-time research engineer in the Natural Language and Conversational AI Lab, Beijing Institute for General Artificial Intelligence, China, from 2024 to present.
\end{IEEEbiography}
\vspace{-250pt}
\begin{IEEEbiography}[{\includegraphics[width=1in,height=1.25in,clip,keepaspectratio]{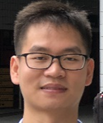}}]{Chao Huang} (S’16-M’17) received his B.Eng. degree in Electrical Engineering and Automation from Harbin Institute of Technology, China, in 2011, M.S. degree in Intelligent Transport System from University of Technology of Compiegne, France, in 2013, and Ph.D. degree in Systems Engineering and Engineering Management from City University of Hong Kong, Hong Kong, in 2017.
He is currently an Associate Professor with the School of Computer and Communication Engineering, University of Science and Technology Beijing (USTB). His research interests include data mining, computational intelligence, and energy informatics.
\end{IEEEbiography}

\EOD

\end{document}